\documentclass[conference]{IEEEtran}

\usepackage[utf8]{inputenc}
\usepackage[T1]{fontenc}
\usepackage{amsmath, amssymb, amsfonts}
\usepackage{graphicx}
\usepackage{subcaption}
\usepackage[font=footnotesize]{caption}
\usepackage{booktabs, array, tabularx, multirow}
\usepackage{float}
\usepackage{enumitem}
\usepackage[colorlinks=true, allcolors=blue]{hyperref}
\usepackage{stfloats}
\usepackage[ruled, linesnumbered]{algorithm2e}
\usepackage{cite}
\usepackage{balance}

\usepackage{tikz}
\usepackage{eso-pic}

\renewcommand{\baselinestretch}{0.952}
\newcommand\PlaceConfHeaderAndCopyright{
  \AddToShipoutPictureFG*{
    
  }
}
\PlaceConfHeaderAndCopyright

\begin{document}

\title{Ensemble Feature Selection and Harris Hawks Optimization for Explainable Mental Health Risk Prediction in Female Sex Workers}

\author{\IEEEauthorblockN{Ahnaf Atef Choudhury}
\IEEEauthorblockA{\textit{Department of Information Sciences and Technology} \\
\textit{George Mason University}\\
USA \\
achoudh9@gmu.edu}
\and
\IEEEauthorblockN{Md. Parvej Hoque Palash}
\IEEEauthorblockA{\textit{Department of Computer Science and Engineering} \\
\textit{Jahangirnagar University}\\
Bangladesh \\
palash.stu2018@juniv.edu}
\and
\IEEEauthorblockN{Shahriar Siddique Ayon}
\IEEEauthorblockA{\textit{Department of Computer Science} \\
\textit{AIUB}\\
Bangladesh \\
shahriarayon63@gmail.com}
\and
\IEEEauthorblockN{Ramkrishna Saha}
\IEEEauthorblockA{\textit{Department of Computer Science} \\
\textit{The University of Texas at Dallas}\\
USA \\
ramkrishna.saha@utdallas.edu}
\and
\IEEEauthorblockN{Abdullah Al Mamun}
\IEEEauthorblockA{\textit{Department of Computer Science and Engineering} \\
\textit{Dhaka University of Engineering and Technology}\\
Bangladesh \\
mamun.duet.bd@gmail.com}
}

\maketitle

\begin{abstract}
One of the significant mental health issues affecting female sex workers (FSWs) is mental disorders, especially depression. Exposure to violence, stigma, and economic hardship further increases their psychological risk. Current machine learning (ML) models are typically ineffective at capturing the high-dimensional and complex risk patterns that exist in this marginalized group. This paper suggests a hybrid predictive model that merges an ensemble feature selection strategy using ANOVA and mutual information and Harris Hawks optimization-tuned logistic regression and represents a new application of swarm intelligence to predict mental health in vulnerable groups. The explainable AI (XAI) methods can be used to understand the factors of trauma associated with model predictions. When applied to a group of 3,005 FSWs, it can be seen that the proposed model is more effective than traditional classifiers, with an accuracy of 95.78\%, an F1 score of 95.77\%, and an AUC of 0.96, and identifying post-traumatic stress, client-related violence, and occupational factors as major contributors to depression. This work bridges the gaps between conventional and ML approaches to develop an XAI tool that enables vulnerable groups to receive early assistance, evidence-based targeted psychosocial care, and health planning.
\end{abstract}


\section{Introduction}

Mental health issues are still a big problem for public health around the world in the 21st century. Prediction of mental health status among female sex workers (FSWs) is still a crucial public health issue given the confluence of structural violence, stigma, economic vulnerability, and occupational risks\cite{Kaya2025}. According to the World Health Organization (WHO), nearly one in seven people worldwide, about 1.1 billion individuals, were living with a mental disorder in 2021, with depression and anxiety being the most prevalent conditions \cite{who2025mental}. Depression affects about 4\% of the global population, including 5.7\% of adults, with higher prevalence among women than men, and together with anxiety disorders leads to an estimated annual loss of nearly US\$1 trillion in productivity \cite{who2025depression}. These disorders are a leading cause of long-term disability worldwide and are expected to rise significantly by 2040 \cite{zhang2025global}. 

Vulnerable groups exposed to overlapping structural, social, and occupational stressors face higher rates of depression, post-traumatic stress disorder (PTSD), anxiety, and suicidality. Among these groups, FSWs are highly marginalized and face severe risks, including violence from partners and clients, stigma, social exclusion, HIV exposure, and economic and housing insecurity \cite{Tutlam2025}. These factors intensify psychological distress and limit access to care, with studies reporting depression rates up to 52.7\% and PTSD rates of 53.6\% among FSWs, far higher than in the general population \cite{nami2025mental}.

Traditional methods has identified these associations, whereas machine learning (ML) has demonstrated significant potential in modeling intricate behavioral and health outcomes, including treatment-seeking behavior, risky sexual behavior, and HIV risk stratification \cite{kebede2023spatial}. Existing techniques frequently depend on traditional statistics or simplistic classifiers that inadequately address high-dimensional, nonlinear, and imbalanced datasets, as well as intricate interactions between exposure to violence, socioeconomic variables, and health indicators \cite{abubakkar2025explainable, bernal2025detection}. Its insufficient hybrid feature selection and sophisticated optimization further deteriorate the levels of accuracy, stability, and interpretability \cite{Kroehn_10.3389, ayon2026explainable}. These constraints highlight the importance of more complex and reliable predictive models to capture the complex mental health risk profile of FSWs and enable early intervention, psychosocial support, and planning of health services.

To address these gaps, this study proposes a hybrid model, which integrates ensemble feature selection with Harris Hawks Optimization (HHO)-tuned Logistic Regression (LR). The algorithm is tried on a community-based dataset, which is already cleaned up. Explainable AI (XAI) LIME also provides understandable, instance-level explanations of model predictions, and, therefore, the decisions made are easier to understand and trust. Major Initiatives of this work:

\begin{itemize}

\item An ensemble feature selection method that involved ANOVA and mutual information to better identify feature relevance as compared to other methods.

\item HHO is the first mental health prediction model to be applied to marginalized populations with high accuracy and HFO-optimized LR, making it better than the baseline models.

\item Implement XAI techniques to demonstrate important traumatic and occupational variables to predict depression, enhancing transparency and clinical applicability.

\end{itemize}

The rest of the paper is structured in the following way: Section \ref{sec:Lr} provides the review of the relevant work, Section \ref{sec:Method} describes the proposed methodology, Section \ref{sec:Result} provides and discusses the results of the experiment, and the last section \ref{sec:Con} provides a conclusion of the study and the future directions of the research.

\section{Literature Review}
\label{sec:Lr}
Interdependent risks, including violence, HIV exposure, and occupational risks, should be modeled to make accurate predictions of FSWs' mental health. Existing stressors such as housing and food insecurity further exacerbate the risk of depression and PTSD \cite{Tomko2023}. Depression and anxiety are also strongly linked to intimate partner violence (IPV) and violence committed by clients, with emotional IPV having the strongest effect \cite{leis2021intimate}. Muhlen et al. assert that stigma and social exclusion contribute to enhancing mental health risks further, which explains why psychosocial interventions are highly necessary \cite{muhlen2023psychische}. Lowering violence and improving psychosocial support are important steps toward helping FSWs with their mental health issues.

Structural vulnerabilities such as violence and economic instability profoundly affect the mental health and HIV risk of FSWs. Jewkes et al. \cite{Jewkes_182211971} through a multi stage, community based cross sectional survey of 3,005 FSWs, reported alarmingly poor mental health outcomes, with 52.7\% experiencing depression and 53.6\% meeting criteria for PTSD. Similarly, Machisa et al. \cite{Machisa_19137913}, using structural equation modeling on a sample of 1,292 participants, found high prevalence rates of binge drinking (50\%), depressive symptoms (43\%), PTSD symptoms (9\%), and suicidal ideation (21\%).

Studies consistently show that FSWs face high rates of depression, PTSD, and suicidality, which can be effectively modeled using advanced ML techniques. Zhang et al. \cite{Zhang2023} evaluated seven ML algorithms—including logistic regression (LR), support vector machines (SVM), random forests (RFC), XGBoost, k-nearest neighbors (KNN), naïve Bayes (NB), and neural networks (NN)—reporting acceptable performance $(ROC > 0.51)$ in predicting risky sexual behaviors, with an overall accuracy of 78\% but limited sensitivity (11\%). Nethi et al. \cite{Nethi2024} found that the light gradient boosting machine achieved the strongest predictive power for identifying candidates suitable for HIV pre-exposure prophylaxis (PrEP), with an area under the curve (AUC) of 0.88. In a similar study, it was showed that decision trees (DT), SVM, and RFC were better at handling SMOTE-processed data and that the accuracy and precision of decision-making processes were 0.871 and 0.960, respectively \cite{He2022}. Qualitative studies emphasise stigma, violence, and inadequate psychosocial support as significant factors contributing to poor mental health among FSWs \cite{Morgan2023}. These structural and psychosocial factors can enhance ML models for more accurate mental health prediction in this population.

The mental health prediction among FSWs depends on such methods as chi-square, ANOVA, mutual information (MI), Boruta, and tree-based models to model intricate risk patterns. Chi-squared tests were applied to examine relationships between categorical variables, such as work conditions and mental health outcomes \cite{Kroehn_10.3389}. The RFC model is the most effective among the four ML models used by Fauste et al. \cite{Ndikumana2025} with a 75\% accuracy of comorbidity of mental diseases and an 88.8\% accuracy of factors that lead to mental health vulnerability.

Integrating ML and XAI enhances the accuracy and interpretability of mental health assessments, making it possible to intervene and support them more effectively. According to Saha et al. \cite{Saha_11136272}, the dynamic ensemble selection methods, including KNORA, have an accuracy of 81\%, which clarifies the importance of using multiple models to enhance predictive accuracy. Masudur et al. \cite{Kanchon2026} used the SMOTE to overcome the class imbalance and used SHapley Additive exPlanations (SHAP) to explain model results. SHAP successfully described feature contributions, and the RFC model performed the best (accuracy = 0.66, recall = 0.69). According to these studies, integrating ML and XAI enhances predictability and interpretability in practical applications related to mental health.

Existing research relies on limited models and struggles with complex risk factors that depend on each other. Its performance is moderate, and it poorly generalizes on imbalanced data. Most studies also lack integrated feature selection, optimization, and XAI, which lowers both accuracy and interpretability. This study addresses these gaps with a unified hybrid ML and XAI framework.

\section{Methodology}
\label{sec:Method}
\begin{figure}[htbp]
    \centering
    \includegraphics[width=\linewidth]{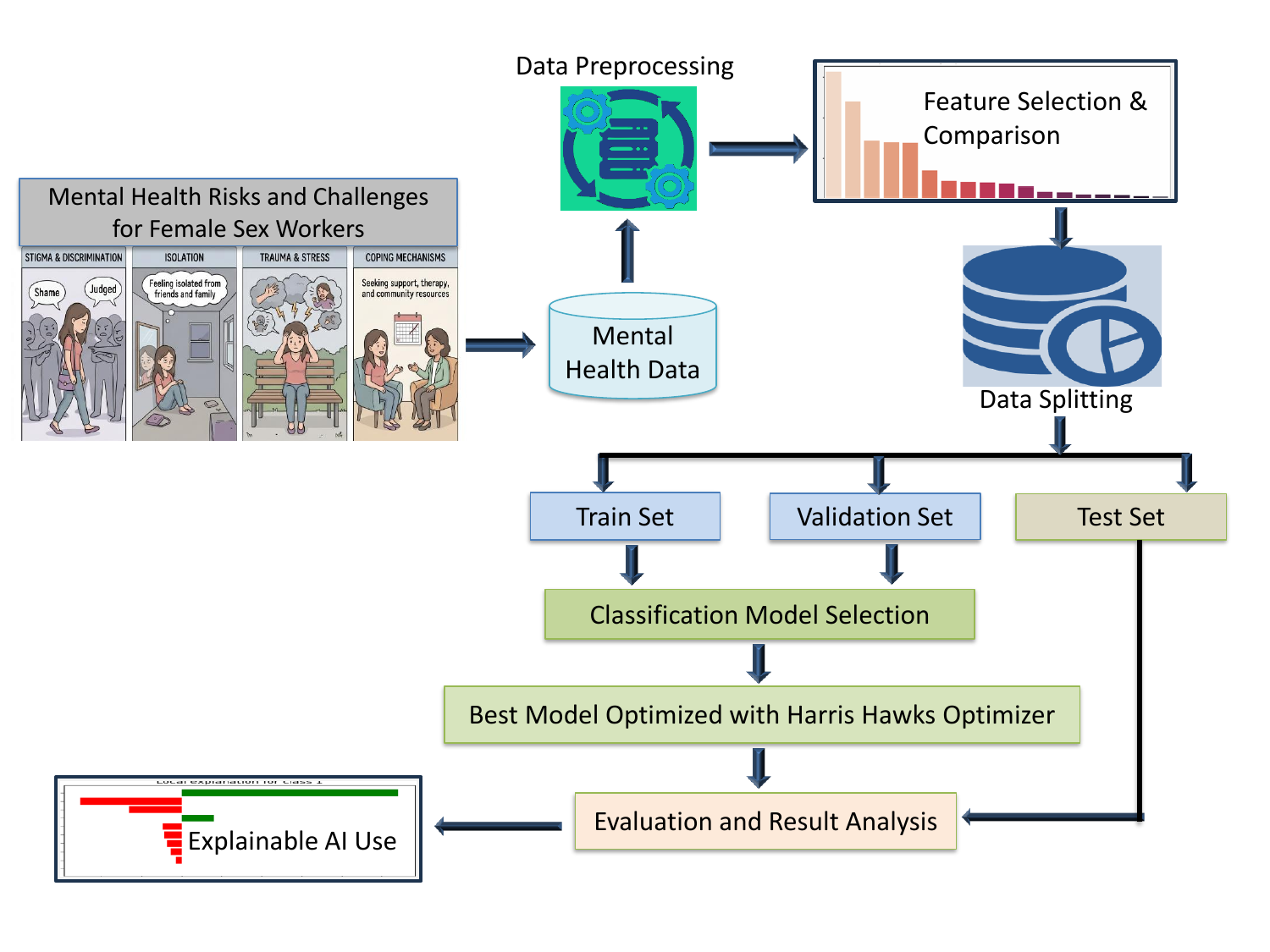}
    \caption{Proposed Methodological Framework for Depression Prediction among Female Sex Workers.}
    \label{fig:proposed}
    \end{figure}

This section is a clear and well-organized presentation of the overall methodological framework of the study. It begins with the data collection, preprocessing, feature selection, and the development of optimized models, and XAI is used to extract important positive and negative features. Fig. \ref{fig:proposed} outlines the suggested framework to predict mental health status among FSWs on the basis of several perspectives.

\subsection{Data Collection, Cleaning, and Preprocessing}
The statistics in this article are based on a giant community-based cross-sectional survey of 3,005 adult FSWs carried out in all nine provinces of South Africa \cite{Milovanovic2021}. The research was aimed at learning important details about their lives, such as HIV status, mental health problems, such as depression and PTSD, violence experience, and work-related issues. The data were gathered in locations where sex worker support programs already exist, and hence it is easier to access the participants by the virtue of having trusted networks. Accurate responses were collected using structured questionnaires and recorded in real time using REDCap. FSWs were also engaged in the entire process, during the survey design and in data collection, which made the information more applicable, trustworthy, and realistic.

The dataset initially had 20 columns with 3,005 rows and seven missing columns. In the numerical columns, Years worked as sex worker, Age of first sex, Number of clients in past day, and Earning potential per client, the missing values were replaced with the mean. Missing values in categorical columns were completely removed. Label encoding of all categorical variables transformed them into numerical form to analyze them using MLS. The column of enrolment date was divided into day, month, and year columns. Superfluous or redundant columns were also eliminated in order to simplify the dataset. The processed final dataset will have 2,911 rows and 19 feature columns.

\subsection{Hybrid Ensemble Feature Selection Strategy}
\begin{figure}[htbp]
    \centering
    \includegraphics[width=\linewidth]{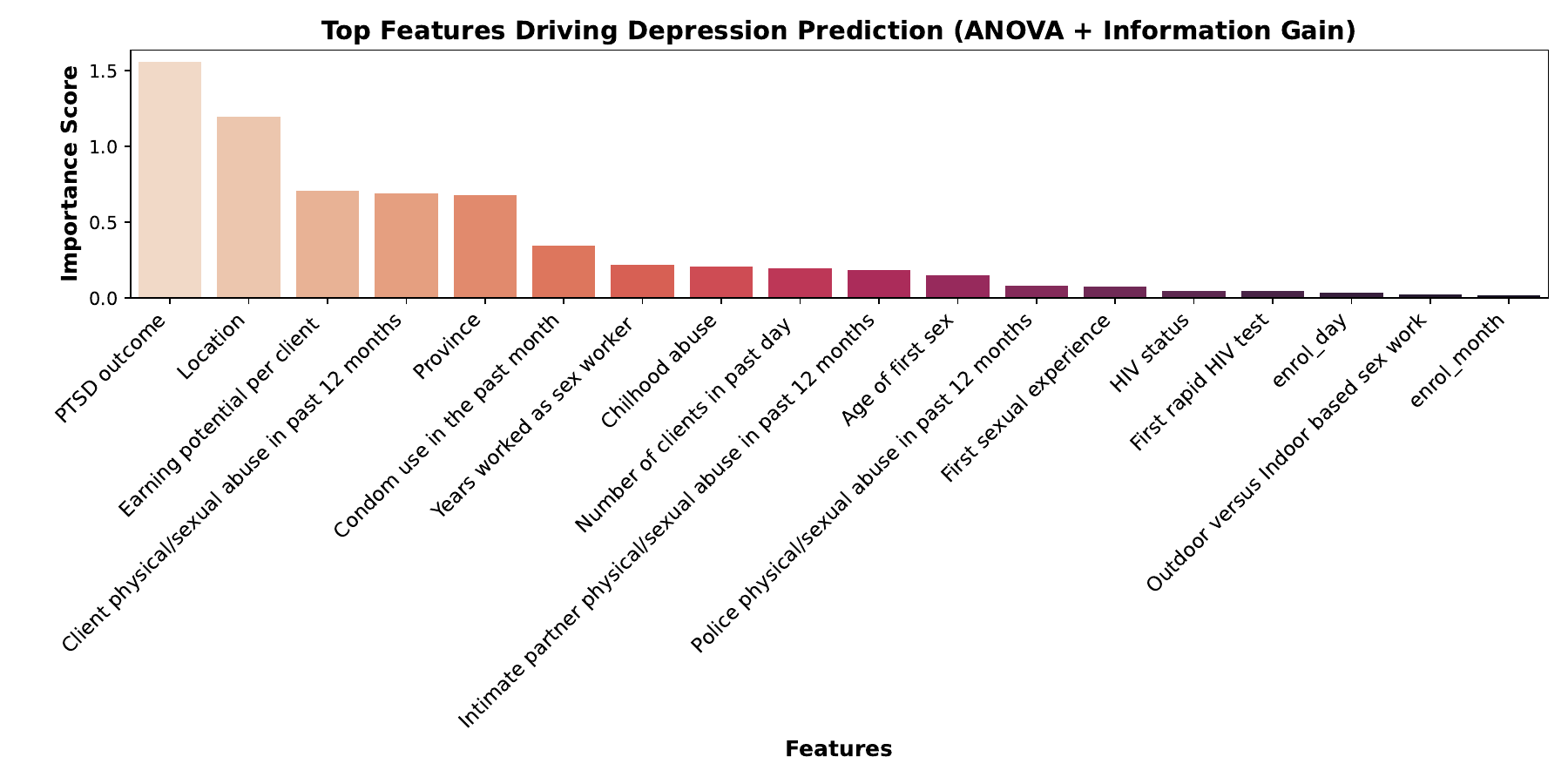}
    \caption{Key Features Driving Depression Prediction Identified by Ensemble Feature Selection.}
    \label{fig:feature_importance}
    \end{figure}

After preparing the data, we using feature selection to determine the most significant variables. We tested a variety of approaches such as chi-squared, ANOVA, MI, Boruta, tree-based algorithms, and ensemble. Among them, there is the ensemble ANOVA and MI technique, which is a combination of statistical analysis of variance and information gain to sturdily detect the most significant features. The ranking of features according to the importance of features using the ensemble feature selection method is shown in figure \ref{fig:feature_importance}.

Figure \ref{fig:feature_importance} presents the ranking of features influencing depression prediction among sex workers, as determined by the ensemble ANOVA and MI method. PTSD outcome was the most important (1.558), followed by Location (1.195) and Earning potential per client (0.710). Other key features included Client physical/sexual abuse (0.688), Province (0.682), and Condom use in the past month (0.347). Other features such as years worked as a sex worker (0.218), childhood abuse (0.206), and number of clients in the past day (0.195) also played a role; the rest of the features had lower scores (0.184-0.019), which means that they had less impact on the depression prediction.

According to the results of the ensemble feature selection, the least significant features were filtered off, resulting in an optimized set of 11 features and 2,911 observations. Location and Province capture regional structural factors (e.g., violence, access to care, and economic disadvantage) not directly observed but relevant in South Africa. They are used only for contextual risk stratification and referral guidance, not as causal individual predictors. The target variable, Depression, consists of 1,530 positive cases and 1,381 negative cases, corresponding to labels 1 and 0, respectively. The data was divided into 80\% training and 20\% testing; 20\% of the training was used as a validation.

\subsection{Model Selection and Hyperparameter Tuning}
We tested several ML and DL models for depression prediction, tuning their hyperparameters for maximum performance. Random Forest Classifier (RFC) is an ensemble of 100 decision trees with max depth 10 and minimum 2 samples per leaf, designed to reduce overfitting. k-Nearest Neighbors (kNN) classifies instances based on the majority label of the 5 nearest neighbors using Euclidean distance. Support Vector Classifier (SVC) separates classes using an RBF kernel with $C = 1.0$ and $\gamma = 0.1$ to find the optimal hyperplane. Light Gradient Boosting Machine (LGBM) is a fast, gradient-boosted tree model using 200 estimators, learning rate of 0.05, and maximum depth of 7. Artificial Neural Network (ANN) with three dense layers (64, 32, 16 neurons), ReLU activation, and Adam optimizer (learning rate 0.001) was used to capture nonlinear relationships. Logistic Regression (LR) models the probability of a binary outcome using a logistic function, configured with L2 regularization, regularization strength $C = 1.0$, and the 'lbfgs' solver.

Among all models, LR performed best, and its performance was further enhanced using advanced swarm intelligence optimization techniques, including Particle Swarm, Ant Colony, Genetic Algorithm, and Harris Hawks Optimization (HHO), with HHO-based LR achieving the highest overall performance. HHO-based LR enhances standard LR by integrating HHO to efficiently search the parameter space and identify optimal model weights. The model estimates the probability of the positive class as:

\begin{equation}
P(y=1|X) = \frac{1}{1 + e^{-(\mathbf{X}\mathbf{w} + b)}}
\end{equation}

Where $\mathbf{w}$ and $b$ are the weights and bias. The HHO algorithm was configured with a population size of 30, maximum iterations of 50, and an escape energy parameter $E_0 = 2$, which together maximized predictive performance and effectively handled class imbalance.

\subsection{Evaluation Metrics for Classification}
In binary classification, the accuracy is the general percentage of correct predictions. Precision is the ratio between the number of predicted positives that are actually positive, whereas recall (or sensitivity) represents the number of true positive cases detected by the model. The F1-score gives an equilibrium between precision and recall, and the AUC gives the capacity of the model to differentiate the two classes, in which the higher the values, the better the discrimination is.

\subsection{Explainable AI LIME}
XAI assists in making ML models understandable, indicating why they make particular predictions. We applied LIME in the study, which clarifies individual predictions by pointing to the contribution of each feature, which offers clear insights on the local level that complement global approaches such as SHAP and SHAPASH \cite{siddique2024explainable}. The LIME explanatory model is represented as follows:

\begin{equation} 
\hat{g}\left( x \right)=arg min_{g\epsilon G} L\left( f,g,\pi_{x\acute{}} \right)+\Omega\left( g \right) 
\end{equation}

In LIME, the complex model $f$ is approximated locally by an interpretable model $\hat{g}(x) \in G$, which minimizes the loss function $L(f, g, \pi_{x'})$ to achieve a balance between fidelity to the original model and interpretability.

\section{Results and Discussion Analysis}
\label{sec:Result}
All models were developed and tested using the free version of Google Colab, providing a simple and efficient platform for experimentation. Models were tuned using 5 fold cross-validation to optimize performance on the training data. We compared model performance before and after applying feature selection and correlation analysis to better understand their effects. To enhance interpretability, LIME was used to identify features that positively or negatively influenced predictions. The overall process and key findings are outlined step by step below.

\subsection{Baseline Model Performance}
Table \ref{tab:model_performance}, the baseline model comparison shows that HHO-based LR achieved the best overall performance with 93.06\% accuracy, 92.82\% precision, 93.20\% recall, 93.01\% F1-score, and 0.94 AUC. Among standard models, LR performed strongest with 92.28\% accuracy and 0.93 AUC, followed by ANN (91.35\%, 0.92 AUC) and LGBM (90.42\%, 0.91 AUC). k-NN showed moderate performance (89.22\%, 0.89 AUC), while RFC (87.85\%, 0.88 AUC) and SVC (87.36\%, 0.87 AUC) achieved comparatively lower results. This highlights the effectiveness of the HHO-based optimization in improving model performance.

\begin{table}[htbp]
\caption{Baseline Model Performance Comparison for Depression Prediction}
\centering
\resizebox{\columnwidth}{!}{
\begin{tabular}{l p{1.cm}p{0.8cm}p{0.8cm}p{1.1cm}c}
\hline
\textbf{Model} & \textbf{Accuracy (\%)} & \textbf{Precision (\%)} & \textbf{Recall (\%)} & \textbf{F1-score (\%)} & \textbf{AUC} \\
\hline
RFC             & 87.85 & 87.62 & 87.94 & 87.78 & 0.88 \\
k-NN            & 89.22 & 89.05 & 89.40 & 89.22 & 0.89 \\
SVC             & 87.36 & 87.15 & 87.50 & 87.32 & 0.87 \\
LGBM            & 90.42 & 90.18 & 90.55 & 90.36 & 0.91 \\
ANN             & 91.35 & 91.10 & 91.55 & 91.32 & 0.92 \\
LR              & 92.28 & 92.05 & 92.50 & 92.27 & 0.93 \\
HHO-LR    & 93.06 & 92.82 & 93.20 & 93.01 & 0.94 \\
\hline
\end{tabular}
\label{tab:model_performance}
}
\end{table}

\subsection{Performance After Feature Selection}
After feature selection (Table \ref{tab:feature_selection_performance}), performance improved across all methods, with the proposed ANOVA + IG achieving the best results. HHO-based LR reached the highest performance with 95.78\% accuracy, 95.60\% precision, 95.95\% recall, 95.77\% F1-score, and 0.96 AUC, followed by LR (94.54\%, 0.95 AUC). In comparison, Boruta and ensemble (MI + RFE) methods showed slightly lower performance, with HHO-based LR achieving up to 94.31\% accuracy and 0.95 AUC. 

\begin{table}[htbp]
\caption{Performance Comparison After Feature Selection}
\centering
\resizebox{\columnwidth}{!}{
\begin{tabular}{p{1.1cm} l p{1.cm}p{0.8cm}p{0.8cm}p{1.cm}c}
\hline
\textbf{Feature Selection} & \textbf{Model} & \textbf{Accuracy (\%)} & \textbf{Precision (\%)} & \textbf{Recall (\%)} & \textbf{F1 score (\%)} & \textbf{AUC} \\
\hline
\multirow{3}{*}{Boruta} 
 & LR              & 93.14 & 92.95 & 93.35 & 93.15 & 0.94 \\
 & ANN             & 92.81 & 92.60 & 93.05 & 92.82 & 0.93 \\
 & HHO-LR    & 94.16 & 93.95 & 94.40 & 94.17 & 0.95 \\
\hline
Ensemble & LR              & 93.22 & 93.05 & 93.40 & 93.22 & 0.94 \\
 (MI+RFE) & XGBoost         & 92.61 & 92.40 & 92.85 & 92.62 & 0.93 \\
  & HHO-LR    & 94.31 & 94.10 & 94.55 & 94.32 & 0.95 \\
\hline 
 Proposed& ANN             & 93.46 & 93.25 & 93.70 & 93.47 & 0.94 \\
 (ANOVA& LR              & 94.54 & 94.35 & 94.75 & 94.55 & 0.95 \\
 + IG) & \textbf{HHO-LR}    & \textbf{95.78} & \textbf{95.60} & \textbf{95.95} & \textbf{95.77} & \textbf{0.96} \\
\hline
\end{tabular}
\label{tab:feature_selection_performance}
}
\end{table}

HHO improved recall by 1.20\% and F1-score by 1.22\% over grid-search LR; while the absolute gain is modest, this corresponds to approximately $\sim 36$ additional correct depression detections in the cohort, justifying its use in high-stakes screening settings where the cost of false negatives outweighs the minimal computational overhead.

\begin{figure}[htbp]
    \centering
    \includegraphics[width=\linewidth]{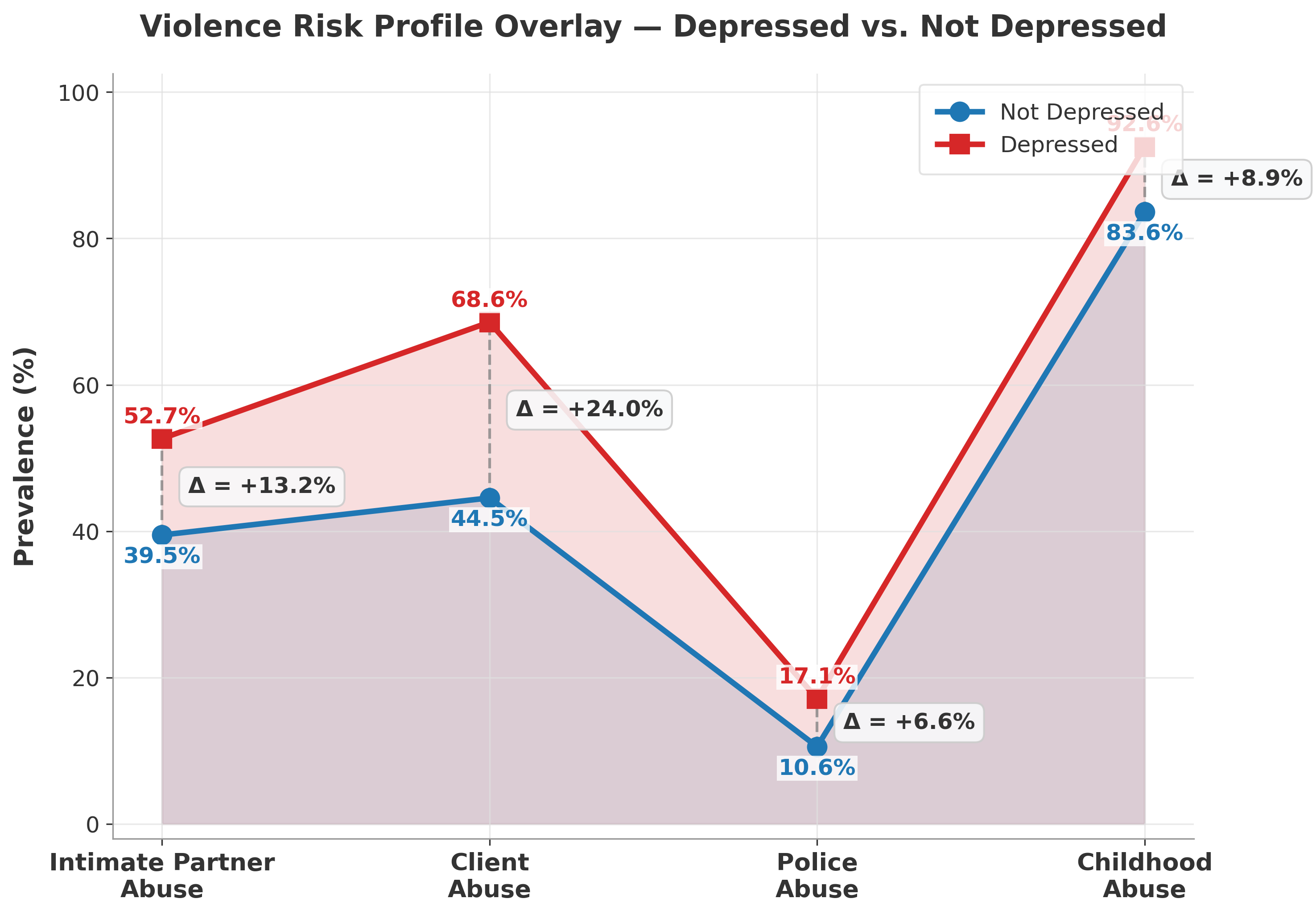}
    \caption{Violence Exposure Prevalence Among Depressed and Non-Depressed Groups.}
    \label{fig:Violence_Exposure}
    \end{figure}

Figure \ref{fig:Violence_Exposure} compares violence exposure between depressed and non-depressed groups, showing consistently higher prevalence among the depressed cohort. Client abuse shows the largest gap (68.6\% vs. 44.5\%, $\Delta = +24.0\%$), followed by intimate partner abuse (52.7\% vs. 39.5\%, $\Delta = +13.2\%$). Childhood abuse remains highest overall (92.6\% vs. 83.6\%, $\Delta = +8.9\%$), while police abuse is lower but still elevated (17.1\% vs. 10.6\%, $\Delta = +6.6\%$). On the whole, these findings suggest that depression is closely related to a greater exposure to various types of violence, especially those related to clients and partners.

\subsection{Explaining Predictions Using LIME}
\begin{figure}[htbp]
    \centering
    \includegraphics[width=\linewidth]{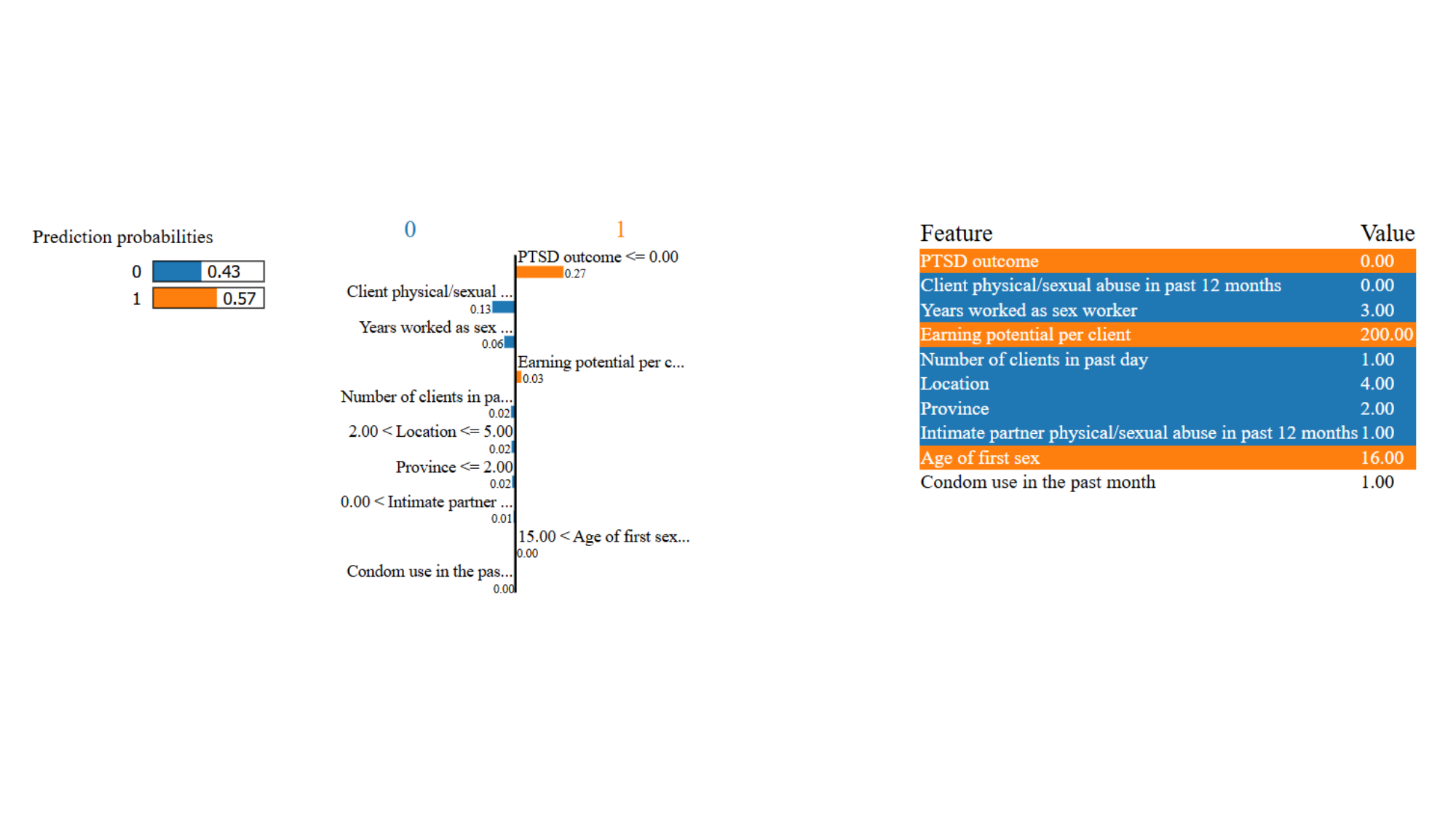}
    \caption{LIME Explanation of Key Features Driving Depression Prediction.}
    \label{fig:LIME_1}
    \end{figure}

We used LIME tabular and feature importance plots to highlight how individual features influenced the model’s predictions. To assess generalizability beyond a single case, LIME explanations were aggregated across all 582 test instances. PTSD outcome and client abuse consistently emerged as the top two predictors in 94.3\% and 87.1\% of cases, respectively, and their importance closely aligned with the ANOVA+MI ranking (Spearman $\rho = 0.91$, $p < 0.001$), indicating strong population-level stability. The Figure \ref{fig:LIME_1} presents a tabular explanation of a binary prediction with LIME that gives more preference to class 1 (0.57 vs. 0.43), which implies that the model predicts higher risk. PTSD outcome (+0.27) is the strongest contributor, followed by recent client abuse (+0.13) and years in sex work (+0.06). Other factors, including earning potential, client number, location, and province, show smaller effects (+0.02–0.03), while the remaining features contribute minimally. Overall, the prediction is primarily driven by trauma-related and occupational exposure factors.

\begin{figure}[htbp]
    \centering
    \includegraphics[width=\linewidth]{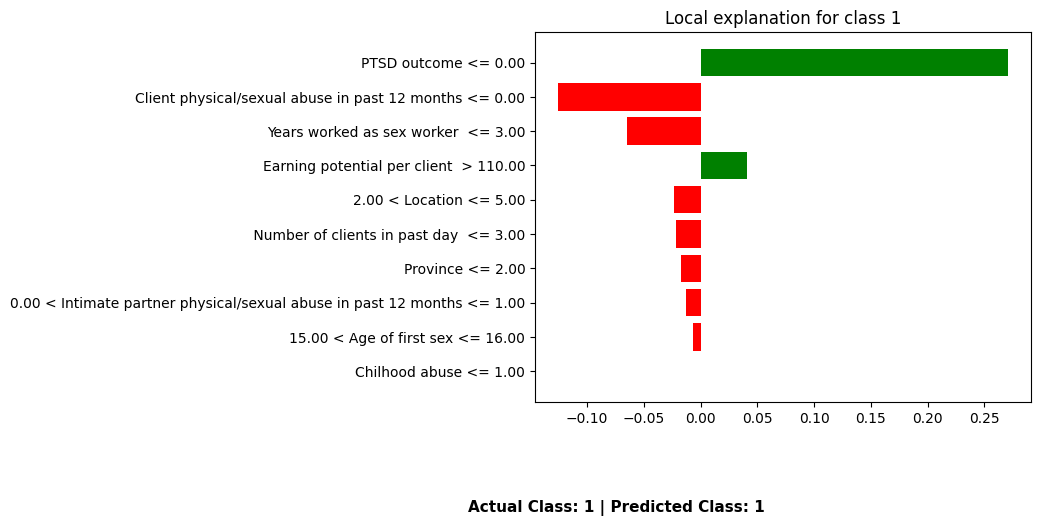}
    \caption{LIME Feature Importance for Correct Depression Prediction.}
    \label{fig:LIME_2}
    \end{figure}

Figure \ref{fig:LIME_2} presents a LIME feature importance plot for a correct prediction (actual = 1, predicted = 1). The top contributors driving the prediction toward class 1 are PTSD outcome $\leq$ 0.00 (+0.27), client abuse in the past 12 months (+0.13), and years worked as a sex worker (+0.06). Moderate positive effects come from earning potential (+0.03), number of clients (+0.02), location (+0.02), and province (+0.02), while intimate partner abuse (+0.01) and features like age of first sex and condom use (~0.00) have minimal impact. Overall, trauma-related and occupational factors dominate, resulting in a confident classification into the higher-risk class.

Past research on mental health prediction for FSWs has shown moderate performance, which is limited by suboptimal feature selection, limited optimization, inability to deal with high dimensional interactions, lack of XAI, and generalization issues \cite{Jewkes_182211971, Nethi2024, Saha_11136272, Kanchon2026}. Our model, on the other hand, has high performance and is mostly led by clinically important factors such as the factors of PTSD and violence exposure. Cross validation assists the internal validity but external generalization is untested. Few differences to the previous work (Zhang et al. \cite{Zhang2023}; Ndikumana et al. \cite{Ndikumana2025}) exist, which is due to the complexity of the datasets and the tasks instead of the superiority of the models. The hybrid model is an ensemble of ensemble feature selection, HHO-optimized LR, and XAI, which enhances both the accuracy and interpretability of mental health predictions. It has a light weight design, so that it can be used on devices without cloud support in real time and on low cost. The model can be integrated into any REDCap-based system to support CHWs in making interpretable depression risk screenings and providing referral support.

\section{Conclusion and Future Work}
\label{sec:Con}
The current study suggested a novel hybrid approach for mental health prediction for FSWs based on the combination of ensemble feature selection and HHO-tuned LR. The model achieved better performance than baseline models and other optimization-based models in mental health datasets through the incorporation of the above techniques: ANOVA, MI and swarm intelligence. Further, LIME increased the interpretability of the findings: they identified important trauma-related and socioeconomic factors that could influence depression in FSWs, which would allow for more explainable and data-based mental health interventions.


Future work will extend the proposed framework through mixed-methods validation, fairness-aware evaluation, multi-country generalization, and real-world deployment. Qualitative follow-up interviews with FSWs will confirm the congruence of the risk factors identified by LIME with the lived experience of FSWs and help to identify other structural determinants for improved modeling. Future studies will test the framework with cohorts from East Africa and South Asia to address the geographic and selection biases of the current dataset from South Africa, and will also include fairness metrics across different subgroups defined by demographic characteristics and exposure to violence. The modular HHO-LR framework can be adapted to other mental health settings, but there is a need for rigorous external validation and benchmarking before claims to broader generalizability can be made. Future studies will also explore the creation of an offline-first mHealth system with encrypted on-device data storage and privacy protection, along with testing in the field to assess usability, cultural acceptability of XAI outputs, and long-term predictive stability in real-world public health environments.


\section*{Data Availability}  
The dataset used in this study is available on Mendeley at the following link:
\url{https://data.mendeley.com/datasets/hfr552s47v/1}

\bibliographystyle{ieeetr}
\bibliography{Ref}
\end{document}